\documentclass[10pt,twocolumn,letterpaper]{article}

\usepackage{iccv}
\usepackage{times}
\usepackage{epsfig}
\usepackage{graphicx}
\usepackage{amsmath}
\usepackage{amssymb}
\usepackage{authblk}

\usepackage{epstopdf, booktabs}
\usepackage{enumitem}
\usepackage{subfigure}
\usepackage[english]{babel}
\usepackage{array}
\usepackage{multirow}

\DeclareMathAlphabet{\mathpzc}{OT1}{pzc}{m}{it} % \mathcal small letters

\usepackage[pagebackref=true,breaklinks=true,letterpaper=true,colorlinks,bookmarks=false]{hyperref}

\iccvfinalcopy % *** Uncomment this line for the final submission

\def\iccvPaperID{23} % *** Enter the ICCV Paper ID here

\ificcvfinal\fi
\begin{document}

%%%%%%%%% TITLE
\title{HydraPlus-Net: Attentive Deep Features for Pedestrian Analysis} 

\author{Xihui Liu$^{1,2*}$,~~~Haiyu Zhao$^{2}$\thanks{X. Liu and H. Zhao share equal contribution.},~~~Maoqing Tian$^2$,~~~Lu Sheng$^1$\\
\vspace{-10pt}
Jing Shao$^{2}$\thanks{J. Shao is the corresponding author.},~~~Shuai Yi$^2$,~~~Junjie Yan$^2$,~~~Xiaogang Wang$^1$\\
{$^1$The Chinese University of Hong Kong~~~~~$^2$SenseTime Group Limited}\\
{\tt\small shaojing@sensetime.com}
}

\maketitle
%\thispagestyle{empty}

%%%%%%%%% ABSTRACT
\begin{abstract}
Pedestrian analysis plays a vital role in intelligent video surveillance and is a key component for security-centric computer vision systems.
Despite that the convolutional neural networks are remarkable in learning discriminative features from images, the learning of comprehensive features of pedestrians for fine-grained tasks remains an open problem.
In this study, we propose a new attention-based deep neural network, named as HydraPlus-Net (HP-net), that multi-directionally feeds the multi-level attention maps to different feature layers.
The attentive deep features learned from the proposed HP-net bring unique advantages:
(1) the model is capable of capturing multiple attentions from low-level to semantic-level, and (2) it explores the multi-scale selectiveness of attentive features to enrich the final feature representations for a pedestrian image.
We demonstrate the effectiveness and generality of the proposed HP-net for pedestrian analysis on two tasks, \ie pedestrian attribute recognition and person re-identification.
Intensive experimental results have been provided to prove that the HP-net outperforms the state-of-the-art methods on various datasets.\footnote{\url{https://github.com/xh-liu/HydraPlus-Net}}
% \vspace{-10pt}
\end{abstract}

%

%%%%%%%%%%%%%%%%%%%%%%%%%%% Introduction %%%%%%%%%%%%%%%%%%%%%%%%%%%
%\vspace{-0.3cm}
\section{Introduction}
\label{sec:intro}
Pedestrian analysis is a long-lasting research topic because of the continuing demands for intelligent video surveillance and psychological social behavior researches.
% Pedestrian analysis plays a vital role in the field of intelligent video surveillance and is a key component for various security-centric computer vision systems.
% For making sense of the vast quantity of visual data collected by large-scale distributed surveillance systems, analysis of pedestrians is essential.
%
Particularly, with the explosion of researches about the deep convolutional neural networks in recent computer vision community, a variety of applications categorized as the pedestrian analysis, \eg pedestrian attribute recognition, person re-identification and \etc, have received remarkable improvements and presented potentialities for practical usage in modern surveillance system.
However, the learning of feature representation for pedestrian images, as the backbone for all those applications, still confronts critical challenges and needs profound studies.

%%%%%%%%%%%%%%%%%%%%% fig: fig1 %%%%%%%%%%%%%%%%%%%%%%%%%%%
\begin{figure}[t]
\centering
\includegraphics[width=0.85\linewidth]{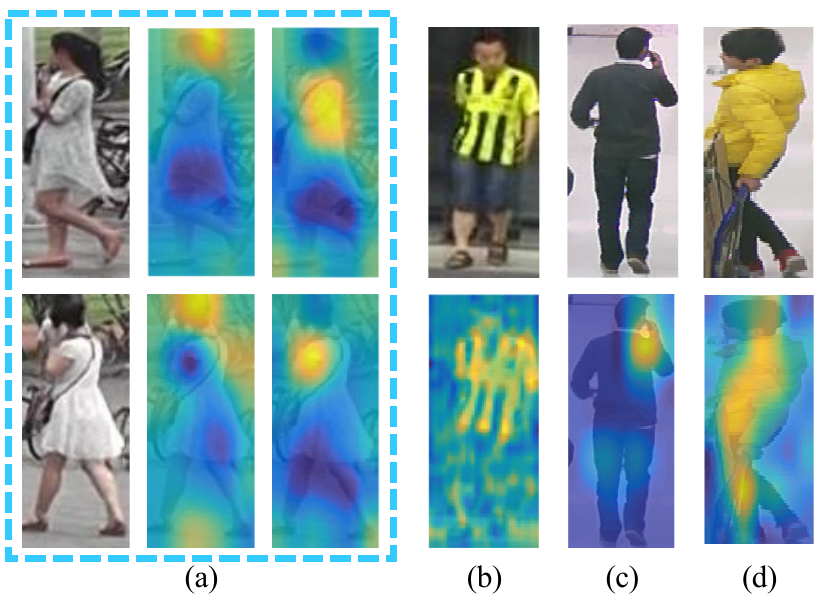}
\caption{
Pedestrian analysis needs a comprehensive feature representation from multi-levels and scales.
% semantic-level (a-b) and low-level (c).
(a) \textit{Semantic-level}: attending features around local regions facilitates distinguishing persons that own similar appearances at a glance, such as ``long hair'' vs. ``short hair'' and ``long-sleeves'' vs. ``short-sleeves''.
(b) \textit{Low-level}: some patterns like ``clothing stride'' can be well captured by low-level features rather than those in high-level.
(c-d) \textit{Scales}: multi-scale attentive features benefit describing person's characteristics, where the small-scale attention map in (c) corresponds to the ``phone'' and a large-scale one in (d) for a global understanding.
%
% (a) Capturing some local characteristics of a pedestrian like ``long hair'' and ``long-sleeves'' can facilitate being distinguished from another appearance-similar person but with ``short hair'' and ``short-sleeves''.
% To distinguish two persons in (a), effectively attending the features around their distinctive local areas like hair and arm is pivotal.
% Effectively attending the effective features of pedestrians is pivotal for pedestrian re-identification in (a).
% Each exemplar is shown with one or two attention maps.
}
\label{fig:fig1}
\vspace{-5pt}
\end{figure}
%%%%%%%%%%%%%%%%%%%%% fig: fig1 %%%%%%%%%%%%%%%%%%%%%%%%%%%

At first, most traditional deep architectures have not extracted the detailed and localized features complementary to the high-level global features, which are especially effective for fine-grained tasks in pedestrian analysis.
For example, it is difficult to distinguish two instances if no semantic features are extracted around hair and shoulders, as shown in Fig.~\ref{fig:fig1}(a).
Also in Fig.~\ref{fig:fig1}(c), the effective features should be located within a small-scale head-shoulder region if we want to detect the attribute ``calling''.
However, existing arts merely extract global features~\cite{li2015multi, sudowe2015person, xiao2016learning} and are hardly effective to location-aware semantic pattern extraction.
Furthermore, it is well-known that multi-level features aid diverse vision tasks~\cite{newell2016stacked,cornia2016deep}.
Similar phenomenon has also happened in the pedestrian analysis, such as the pattern ``clothing stride'' shown in Fig.~\ref{fig:fig1}(b) should be inferred from low-level features, while the attribute ``gender'' in Fig.~\ref{fig:fig1}(d) is judged by semantic understanding of the whole pedestrian image.
Unlike previous approaches that mainly generate the global feature representations, the proposed feature representation encodes multiple levels of feature patterns as well as a mixture of global and local information, and thus it owns a potential capability for multi-level pedestrian attribute recognition and person re-identification.

Facing the drawbacks of recent methods for pedestrian analysis, we try to tackle the general feature learning paradigm for pedestrian analysis by a multi-directional network, called HydraPlus-Net, which is proposed to better exploit the global and local contents with multi-level feature fusion of a single pedestrian image.
%
% Specifically, we propose a multi-directional attention (MDA) module that aggregates multiple level feature maps within the detected spatial regions by the attention maps from different layers in the network.
% %
% Since the attention maps are extracted from different semantic levels, they naturally abstract different levels of visual patterns of the same pedestrian image.
% %
% Moreover, filtering multiple levels of features by the same attention map results in an effective aggregation of multi-level feature fusion from a certain local attention distribution.
% %
Specifically, we propose a multi-directional attention (MDA) module that aggregates multiple feature layers within the attentive regions extracted from multiple layers in the network.
Since the attention maps are extracted from different semantic layers, they naturally abstract different levels of visual patterns of the same pedestrian image.
Moreover, filtering multiple levels of features by the same attention map results in an effective fusion of multi-level features from a certain local attention distribution.
After applying the MDA to different layers of the network, the multi-level attentive features are fused together to form the final feature representation.

The proposed framework is evaluated on two representatives among the pedestrian analysis tasks, \ie pedestrian attribute recognition and person re-identification (ReID), in which attribute recognition focuses on assigning a set of attribute labels to each pedestrian image while ReID aims to associate the images of one person across multiple cameras and/or temporal shots.
Although pedestrian attribute recognition and ReID pay attention to different aspects of the input pedestrian image, these two tasks can be solved by learning a similar feature representation, since they are inherently correlated with similar semantic features and the success of one task will improve the performance of the other.
Compared with existing approaches, our framework achieves the state-of-the-art performance on most datasets.

The contributions of this work are three-fold:

(1) A HydraPlus Network (HP-net) is proposed with the novel multi-directional attention modules to train multi-level and multi-scale attention-strengthened features for fine-grained tasks of pedestrian analysis.
%
% In this way, the local detail information from feature maps of different stages can be better used.

(2) The HP-net is comprehensively evaluated on pedestrian attribute recognition and person re-identification.
State-of-the-art performances have been achieved with significant improvements against the prior methods.

(3) A new large-scale pedestrian attribute dataset (PA-$100$K dataset) is collected with the most diverse scenes and the largest number of samples and instances up-to-date.
The PA-$100$K dataset is more informative than the previous collections and helpful for various pedestrian analysis tasks.

%%%%%%%%%%%%%%%%%%%%% fig: framework %%%%%%%%%%%%%%%%%%%%%%%%%%%
\begin{figure}[t]
\centering
\includegraphics[width=1\linewidth]{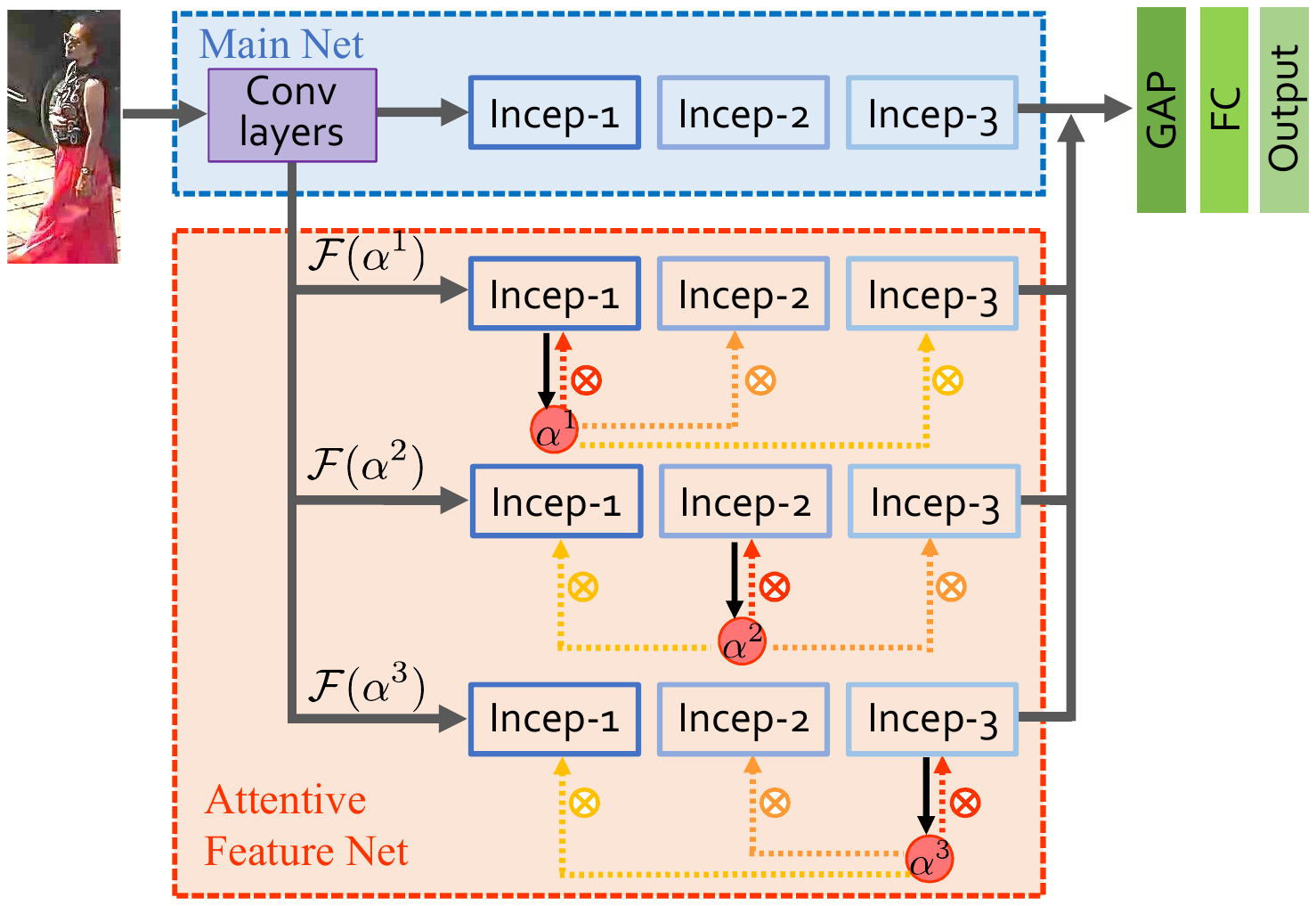}
\caption{A deep HP-Net with a Main Net (M-net) and an Attentive Feature Net (AF-net). The AF-net comprises three multi-directional attention (MDA) modules (\ie~$\mathcal{F}(\alpha^i), i\in\Omega$). Each MDA module includes two components: (1) attention map generation with black solid lines, and (2) attentive features by masking the attention map to different levels of features in hot dash lines. A global average pooling and one fully-connected layer are applied to the concatenated features obtained from the M-net and AF-net.
}
\label{fig:framework}
% \vspace{-0.1cm}
\end{figure}
%%%%%%%%%%%%%%%%%%%%% fig: framework %%%%%%%%%%%%%%%%%%%%%%%%%%%

% \vspace{-0.7cm}

%%%%%%%%%%%%%%%%%%%%%%%%%%% Related work %%%%%%%%%%%%%%%%%%%%%%%%%%%
\section{Related Works}
\label{sec:related_work}
% \vspace{-0.2cm}

\noindent \textbf{Attention models}
% Attention models have achieved great success in neural language processing (NLP) and computer vision.
In computer vision, attention models have been used in tasks such as image caption generation~\cite{xu2015show}, visual question answering~\cite{lu2016hierarchical,xu2016ask} and object detection~\cite{ba2014multiple}.
Mnih~\etal~\cite{mnih2014recurrent} and Xiao~\etal~\cite{xiao2015application} explored hard attention, in which the network attends to a certain region of the image or feature map.
Compared to non-differentiable hard attention trained by reinforce algorithms~\cite{williams1992simple}, soft attention which weights the feature maps is differentiable and can be trained by back propagation.
% soft attention, which generates attention maps to indicate weights of features, is differentiable and can be trained end to end. Yang~\etal~\cite{yang2016stacked} adopted soft attention to guide the network attend on regions related to a certain question.
Chen~\etal~\cite{chen2016attention} introduced an attention to the multi-scale features, and Zagoruyko~\etal~\cite{zagoruyko2016paying} exploited attention in knowledge transfer. In this work, we design a multi-directional attention network for better pedestrian feature representation and apply it to both pedestrian attribute recognition and re-identification tasks. To the best of our knowledge, this is the first work to adopt attention idea in the aforementioned two tasks.

% \vspace{-0.2cm}
\noindent　\textbf{Pedestrian attribute recognition}
Pedestrian attribute has been an important research topic recently, due to its prospective application in video surveillance systems. Convolutional neural networks have achieved great success in pedestrian attribute recognition. Sudowe~\etal~\cite{sudowe2015person} and Li~\etal~\cite{li2015multi} proposed that jointly training multiple attributes can improve the performance of attribute recognition.
Previous work also investigated the effectiveness of utilizing pose and body part information in attribute recognition.
Zhang~\etal~\cite{zhang2014panda} proposed a pose aligned network to capture the pose-normalized appearance differences.
Different from previous works, we propose an attention structure which can attend to important areas and align body parts without prior knowledge on body parts or poselets.

% \vspace{-0.2cm}
\noindent \textbf{Person re-identification}
Feature extraction and metric learning~\cite{koestinger2012large,liao2015person} are two main components for person re-identification.
The success of deep learning in image classification inspired lots of studies on person ReID~\cite{cheng2016person, li2014deepreid, xiao2016learning, wu2016personnet, varior2016gated, su2016deep, ustinova2015multiregion, li2017person, xiao2017joint}.
The filter pairing neural network (FPNN) proposed by Li~\etal~\cite{li2014deepreid} jointly handles misalignment, transforms, occlusions and background clutters.
Cheng~\etal~\cite{cheng2016person} presented a multi-channel parts-based CNN to learn body features from the input image.
% As for metric learning, Koestinger~\etal~\cite{koestinger2012large} introduced an effective strategy to learn a distance metric from equivalence constraints.
% Cross-view Quadratic Discriminant Analysis (XQDA) was proposed by Liao~\etal~\cite{liao2015person} for subspace modeling and metric learning.
In this paper, we mainly target on feature extraction and cosine distance is directly adopted for metric learning.
Moreover, attention masks are utilized in our pipeline to locate discriminative regions which can better describe each individual.

%%%%%%%%%%%%%%%%%%%%% fig: attention_module %%%%%%%%%%%%%%%%%%%%%%%%%%%
\begin{figure}[t]
\centering
\includegraphics[width=0.8\linewidth]{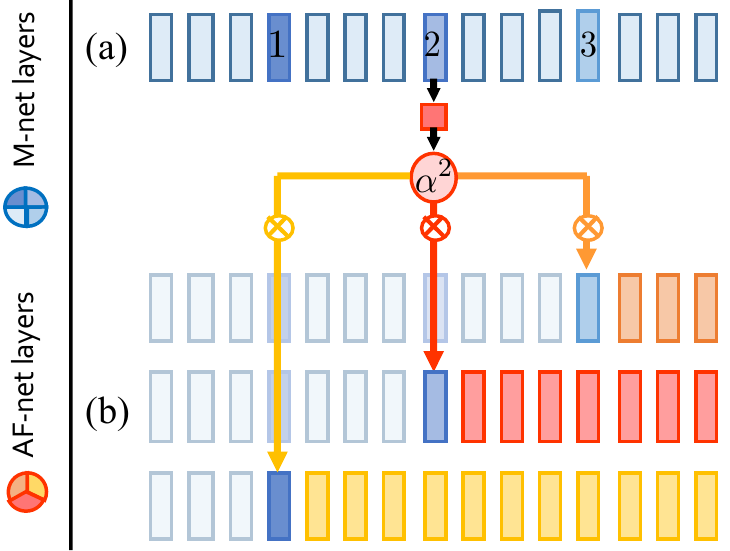}
\caption{An example of the multi-directional attention (MDA) module $\mathcal{F}(\alpha^2)$. The M-net in (a) is presented with simplified layers (indexed by $i\in\Omega$) representing the output layers of three \texttt{inception} blocks. The attention map $\alpha^2$ is generated from block $2$ and multi-directionally masks three adjacent blocks.
}
\label{fig:attention_module}
% \vspace{-0.1cm}
\end{figure}
%%%%%%%%%%%%%%%%%%%%% fig: attention_module %%%%%%%%%%%%%%%%%%%%%%%%%%%

% \vspace{-0.1cm}
%%%%%%%%%%%%%%%%%%%%% HyperPlus Deep Architecture %%%%%%%%%%%%%%%%%%%%%%%%%%
\section{HydraPlus-Net Architecture}
\label{sec:multi_attention}
\vspace{-0.1cm}
% In this paper, we propose a new {\color{red}end-to-end} model named as \textit{HyperPlus-Net} (HP-net).
%
The design of the \textbf{HydraPlus network}\footnote{“Hydra” is a water monster with nine heads. In this work, the network consists of a 9-branch in AF-net (3 MDA modules and 3 attention sub-branches for each MDA), plus an M-net, so it is called HydraPlus Net.} (HP-net) is motivated by the necessity to extract multi-scale features from multiple levels, so as not only to capture both global and local contents of the input image but also assemble its features with different levels of semantics.
As shown in Fig.~\ref{fig:framework}, the HP-net consists of two parts, one is the \textit{Main Net} (M-net) that is a plain CNN architecture, the other is the \textit{Attentive Feature Net} (AF-net) including multiple branches of multi-directional attention (MDA) modules applied to different semantic feature levels.
The AF-net shares the same basic convolution architectures as the M-net except the added MDA modules.
Their outputs are concatenated and then fused by global average pooling (GAP) and fully connected (FC) layers. The final output can be projected as the attribute logits for attribute recognition or feature vectors for re-identification.
In principle, any kind of CNN structure can be applied to construct the HP-net.
But in our implementation, we design a new end-to-end model based on \texttt{inception\_v2} architecture~\cite{ioffe2015batch} because of its excellent performance in general image-related recognition tasks.
As sketched in Fig.~\ref{fig:framework}, each network of the proposed framework contains several low-level convolutional layers and is followed by three \texttt{inception} blocks.
%
% This model seems simple but is non-trivial as it achieves all required abilities and brings them together to boost the recognition capability for the pedestrian analysis.
%
This model seems simple but is non trivial as it achieves all required abilities and brings them together to boost the recognition capability.
% The details and analyses of the MDA modules and the training procedure will be presented in the Section~\ref{sec:ablation_study}.
% %
% Before that, we first provide the architecture of the AF-net and the training strategy for the complete HP-net.

%%%%%%%%%%%%%%%%%%%%% fig: attention_level %%%%%%%%%%%%%%%%%%%%%%%%%%%
\begin{figure}[t]
\centering
\includegraphics[width=\linewidth]{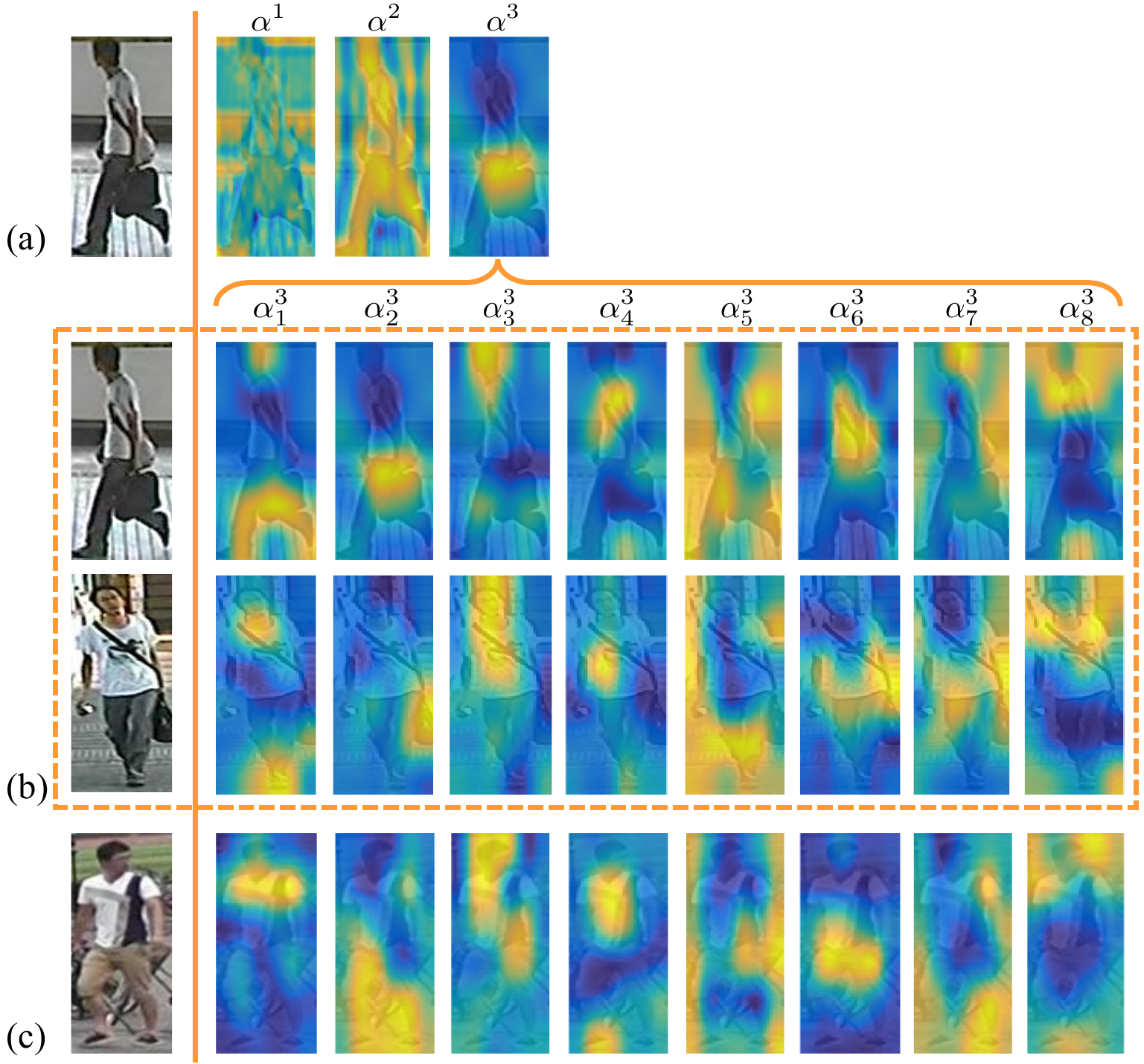}
\caption{The diversity and semantic selectiveness of attention maps. (a) The attention maps generated from three adjacent blocks respond to visual patterns in different scales and levels, in which $\alpha^3$ owns the power to highlight semantic patterns in object level. (b-c) Different channels in attention maps capture different visual patterns related to body parts, salient objects and background.
}
\label{fig:attention_level}
% \vspace{-0.1cm}
\end{figure}
%%%%%%%%%%%%%%%%%%%%% fig: attention_level %%%%%%%%%%%%%%%%%%%%%%%%%%%

\vspace{-0.1cm}
\subsection{Attentive Feature Network}
\label{subsec:model_single_branch}

The Attentive Feature Network (AF-net) in Fig.~\ref{fig:framework} comprises three branches of sub-networks augmented by the multi-directional attention (MDA) modules, namely $\mathcal{F}(\alpha^i), i\in\Omega=\{1,2,3\}$, where $\alpha^i$ are the attention maps generated from the output features of the \texttt{inception} block $i$ marked by black solid lines, and are applied to the output of the $k^\text{th}$ block ($k\in\Omega=\{1,2,3\}$) in hot dash lines.
For each MDA module, there is one link of attention generation and three links for attentive feature construction. Different MDA modules have their attention maps generated from different inception blocks and then been multiplied to feature maps of different levels to produce multi-level attentive features. An example of a MDA module $\mathcal{F}(\alpha^2)$ is shown in Fig.~\ref{fig:attention_module}.
%
% We use $i_k$ to denote the input and output links.
%
The main stream network of each AF-net branch is initialized exactly as the M-net, and thus the attention maps approximately distill similar features as what the M-net extracts.

It is well known that the attention maps learned from different blocks vary in scale and detailed structure.
For example, the attention maps from higher blocks (\eg $\alpha^3$) tend to be coarser but usually figure out the semantic regions like $\alpha^3$ highlights the handbag in Fig.~\ref{fig:attention_level}(a).
But those from lower blocks (\eg $\alpha^1$) often respond to local feature patterns and can catch detailed local information like edges and textures, just as the examples visualized in Fig.~\ref{fig:attention_level}(a).
Therefore, if fusing the multi-level attentive features by MDA modules, we enable the output features to gather information across different levels of semantics, thus offering more selective representations.
Moreover, the MDA module also differs from the traditional attention-based models~\cite{newell2016stacked,xu2015show} that push the attention map back to the \text{same} block, and it extends this mechanism by applying the attention maps to \textit{adjacent} blocks, as shown in lines with varying hot colors in Fig.~\ref{fig:attention_module}.
Applying one single attention map to multiple blocks naturally let the fused features encode multi-level information within the same spatial distribution which is illustrated in Section~\ref{subsec:multi_directional_attentive_features}.

More specifically, for a given \texttt{inception} block $i$, its output feature map is denoted as $\mathbf{F}^i \in \mathbb{R}^{C \times H \times W}$ with the width $W$, height $H$ and $C$ channels.
The attention map $\alpha^i$ is generated from $\mathbf{F}^i$ by a $1\times1$ \texttt{conv} layer with \texttt{BN} and \texttt{ReLU} activation function afterwards, noted as
\begin{equation}
\label{eq:attention_map}
\vspace{-1pt}
\alpha^i = g_\mathrm{att}(\mathbf{F}^i; \boldsymbol\theta^i_{\mathrm{att}}) \in \mathbb{R}^{L\times H \times W},
\vspace{-1pt}
\end{equation}
where $L$ means the channels of the attention map.
In this paper, we fix $L=8$ for both tasks.
And the attentive feature map to the \texttt{inception} block $k$ is an element-wise multiplication
\begin{equation}
\label{eq:attentive_features}
\vspace{-1pt}
\tilde{\mathbf{F}}^{i,k}_l = \alpha^i_l \circ \mathbf{F}^{k}, ~~l\in\lbrace1,\ldots,L\rbrace.
\vspace{-1pt}
\end{equation}
Each attentive feature map $\tilde{\mathbf{F}}^{i,k}_l$ is then passed through the following blocks thereafter, and at the end of MDA module we concatenate the $L$ attentive feature maps as the final feature representation.
We visualized the detailed structure of an MDA module $\mathcal{F}(\alpha^2)$ in Fig.~\ref{fig:attention_module}.
$\alpha^2$ is generated from the \texttt{inception} block $2$ and then applied to feature maps indexed by $k\in\Omega=\{1,2,3\}$, as shown in Fig.~\ref{fig:attention_module}(b).
Note that we prefer the \texttt{ReLU} activation function rather than the \texttt{sigmoid} function to constrain the attention maps so that the attentive regions receive more weights, and the contrast of the attention map is enlarged.
More examples and analyses are shown in Sec.~\ref{sec:ablation_study} to illustrate the MDA's effectiveness.

\subsection{HP-Net Stage-wise Training}
\label{subsec:stage_wise_training}

We train the HP-net in a stage-wise fashion.
Initially, a plain M-net is trained to learn the fundamental pedestrian features.
Then the M-net is duplicated three times to construct the AF-net with adjacent MDA modules, each of which following the framework shown in Fig.~\ref{fig:attention_module}.
%
% Three MDA modules are firstly trained \textit{separately} by fixing all the previous layers before the first block and fine-tuning the rest network parameters and the attention module.
% Three MDA modules are firstly trained \textit{separately} by fixing all the previous layers before the block generating attention maps and fine-tuning the rest network parameters.
% Three MDA modules are first trained \textit{separately} and then joint
%
Since each MDA module consists of three branches where the attention map masks adjacent \texttt{inception} blocks, thus in each branch we only fine-tune the blocks after the attention-operated block.
%
% Finally, we fix both the M-net and AF-net, and train the remaining \texttt{GAP} and \texttt{FC} layers.
After separately fine-tuning three MDA modules in AF-net, we fix both the M-net and AF-net and train the remaining \texttt{GAP} and \texttt{FC} layers.
The output layer to minimize losses defined by different tasks, in which the cross-entropy loss $\mathcal{L}_\text{att}$ is applied for pedestrian attribute recognition, and softmax loss for person re-identification.

%%%%%%%%%%%%%%%%%%%%% fig: attention_mask_feat %%%%%%%%%%%%%%%%%%%%%%%%%%%
\begin{figure}[t]
\centering
\includegraphics[width=1\linewidth]{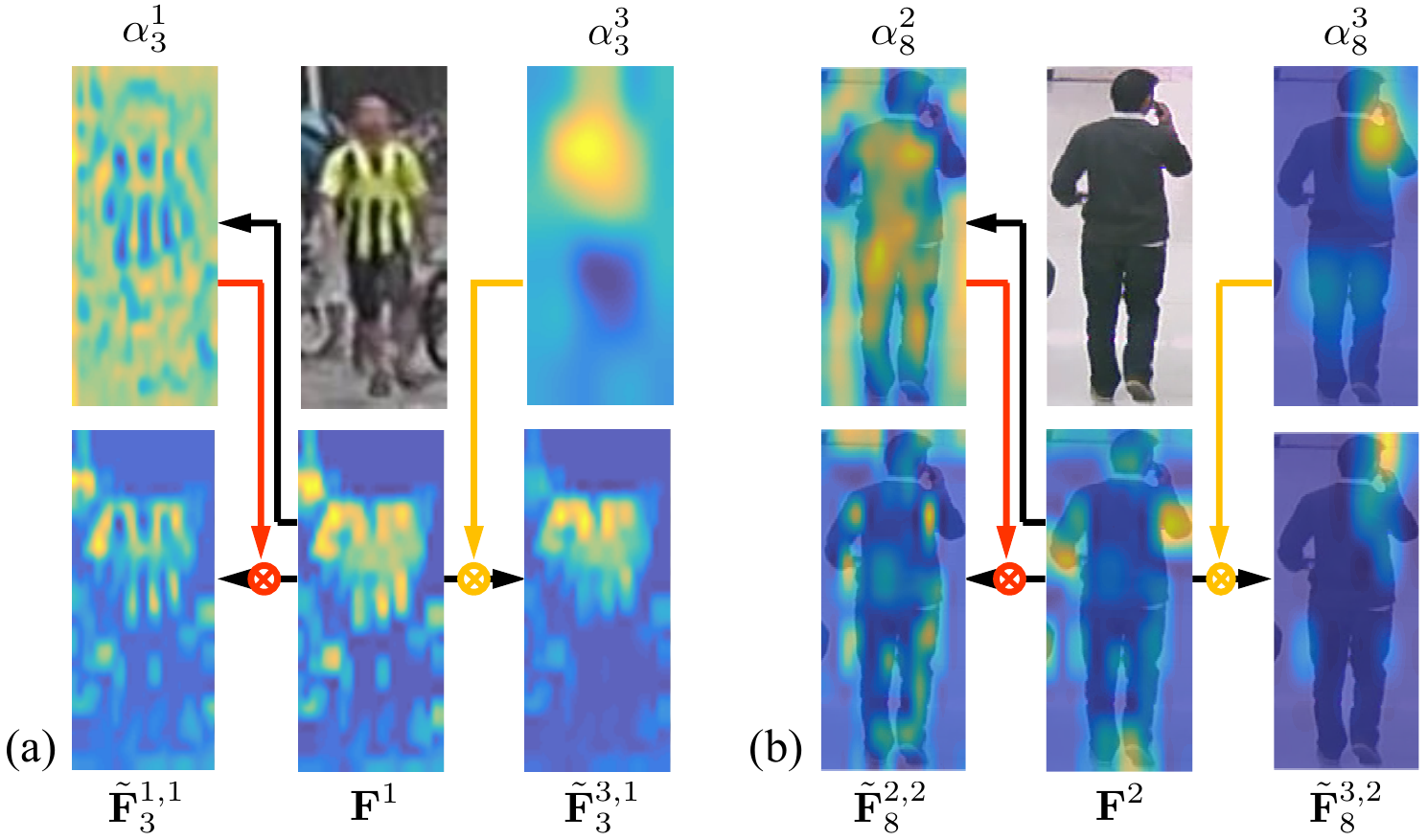}
\caption{
Examples of multi-directional attentive features.
(a) The identification of low-level attributes like ``upper-clothing pattern'' requires the low-level attention connections, for example, applying $\alpha^1_3$ to extract $\tilde{\mathbf{F}}^{1,1}_3$ indicating textures onto the T-shirt. (b) But the semantic or object-level attributes like ``phone'' require high-level attention connections such as applying $\alpha^3_8$ to extract $\tilde{\mathbf{F}}^{3,2}_8$ for the detection of the phone near the ear.
}
\label{fig:attention_mask_feat}
% \vspace{-0.1cm}
\end{figure}
%%%%%%%%%%%%%%%%%%%%% fig: attention_mask_feat %%%%%%%%%%%%%%%%%%%%%%%%%%%

%%%%%%%%%%%%%%%%%%%%% fig: AF-net_analysis %%%%%%%%%%%%%%%%%%%%%%%%%%%
\begin{figure*}[t]
\centering
\includegraphics[width=1\linewidth]{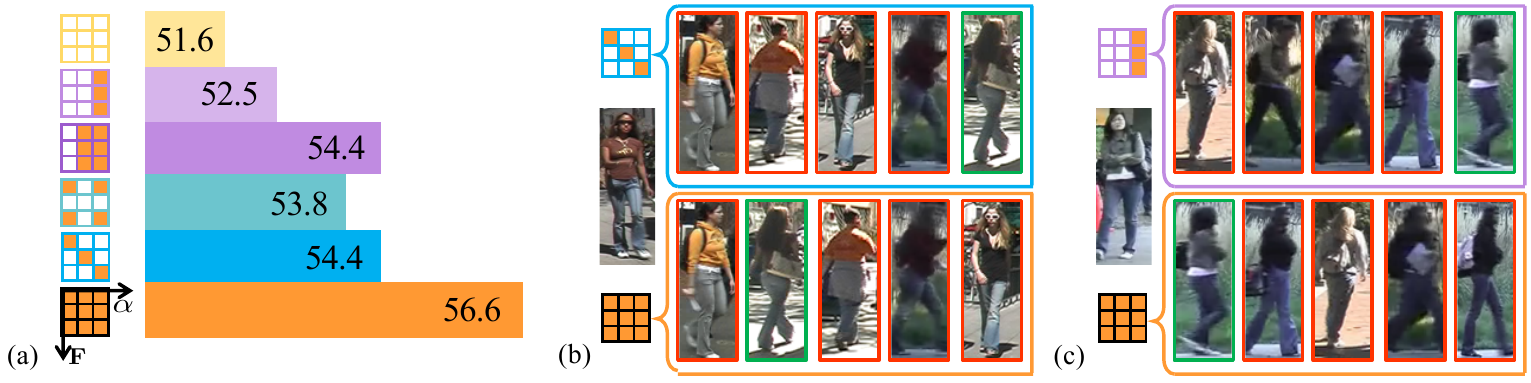}
\caption{Results of discarding partial attention modules or connections compared with that of the complete network fed with all MDA modules on VIPeR dataset. The $3\times 3$ boxes in (a) indicates the indices of different attention maps and their mask directions. The hollow white in each box means the corresponding attentions or directional links have been cut down. Bars are plot by the Top-1 accuracy. (b) and (c) present the qualitative results by the complete network compared with two kinds of partial networks in (a). For a query image shown in the middle, Top-5 results are shown aside with the correct marked by green and the false alarm are red. Best viewed in color.
}
\label{fig:AF-net_analysis}
% \vspace{-0.1cm}
\end{figure*}
%%%%%%%%%%%%%%%%%%%%% fig: AF-net_analysis %%%%%%%%%%%%%%%%%%%%%%%%%%%

%%%%%%%%%%%%%%%%%%%%% Ablation Study on Attended Deep Features%%%%%%%%%%%%%%%%%%%%%%%%%%
\section{Ablation Study On Attentive Deep Features}
\label{sec:ablation_study}

The advantages of HP-net are its capability of learning both multi-level attentions and multi-scale attentive features for a comprehensive feature representation of a pedestrian image.
To better understand these advantages, we analyze the effectiveness of each component in the network with qualitative visualization and quantitative comparisons.

\subsection{Multi-level Attention Maps}
\label{subsec:multi_level_attention_maps}

\noindent\textbf{The level of attention maps.}
The compared exemplars of attention maps from three layers (\ie the outputs of the \texttt{inception} blocks $i\in\Omega=\{1,2,3\}$) are shown in Fig.~\ref{fig:attention_level}(a).
We observe that the attention map from earlier layer $i=1$ prefers grasping low-level patterns like edges or textures, while those from higher layers $i=2$ or $3$ are more likely to capture semantic visual patterns corresponding to a specific object (\eg handbag) or human identity.

\noindent\textbf{The quantity of attention maps.}
Most previous studies~\cite{xu2015show,newell2016stacked} merely demonstrated the effectiveness of an attention-based model with a limited number of channels (\ie, $L=1$ or $2$).
In this study, we explore the potential performance of an attention model with increasing channels in both diversity and consistency.

\textit{1) Attention Diversity.}
Fig.~\ref{fig:attention_level}(b) shows two images of one single pedestrian captured by two cameras, alongside with $L=8$ attention channels of $\alpha^3$ are presented.
From the raw image, it is hard to distinguish these images due to the large intra-class variations from cluttered background, varying illumination, viewpoint changes and \etc
Nevertheless, benefited from the discriminative localization ability of multiple attention channels from one level, the entire features can be captured separately with respect to different attentive areas.
Compared to a single attention channel, the diversity of multiple attention channels enriches the feature representations and improves the chance to accurately analyze both the attributes and identity of one pedestrian.

\textit{2) Attention Consistency.}
We also observe that one attention map generated upon different input samples might be similarly distributed in spatial domain since they highlight the same semantic parts of a pedestrian.
Notwithstanding different pedestrians, shown in Fig.~\ref{fig:attention_level}(b-c), their attention channels $\alpha_3^3$ capture the head-shoulder regions and the channels $\alpha^3_5$ infer the background area.
Since the consistent attention maps are usually linked to salient objects, the selectiveness of these attention maps is thus essential on identifying the pedestrian.

\subsection{Multi-Directional Attentive Features}
\label{subsec:multi_directional_attentive_features}

Apart from the benefits of the multi-level attention maps, the effectiveness of the proposed method also lies on the novel transition scheme.
For instance, the pedestrian in Fig.~\ref{fig:attention_mask_feat}(b) holds a phone near the right ear that cannot be directly captured neither by the feature map $\mathbf{F}^2$ in a lower layer $i=2$, nor by the na\"ive attentive feature maps $\tilde{\mathbf{F}}^{2,2}_8$.
Surprisingly, with the help of a higher level attention map $\alpha^3_8$, the attentive feature map $\tilde{\mathbf{F}}^{2,3}_8$ can precisely attend the region around the phone.
On the other hand, the high-level attention map $\alpha^3_3$ might not be able to capture lower-level visual patterns related to attributes like ``upper-clothing pattern''.
For example, the attention map $\alpha^3_3$ shown in Fig.~\ref{fig:attention_mask_feat}(a) does not point out the local patterns onto the T-shirt, while on the contrary, the low-level attention map $\alpha^1_3$ filters out $\tilde{\mathbf{F}}^{1,1}_3$ that typically reflects these texture patterns.

%%%%%%%%%%%%%%%%%%%%% fig: comp_bar_rap %%%%%%%%%%%%%%%%%%%%%%%%%%%
\begin{figure*}[t]
\centering
\includegraphics[width=1\linewidth]{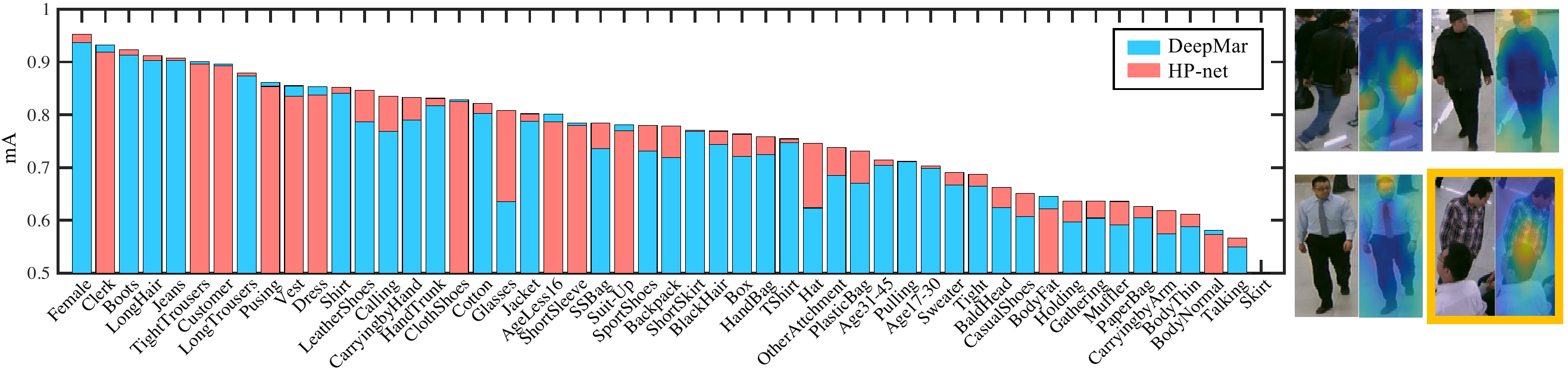}
\caption{
Mean accuracy scores for all attributes of RAP dataset by HP-net and DeepMar marked with red and blue bars respectively. The bars are sorted according to the larger mAs between two methods. The HP-net outperforms DeepMar especially on ``glasses'' and ``hat'' which have the exemplar sample listed aside. The sample in orange provides a failure case of predicting the attribute ``talking''.
}
\label{fig:comp_bar_rap}
% \vspace{-0.1cm}
\end{figure*}
%%%%%%%%%%%%%%%%%%%%% fig: comp_bar_rap %%%%%%%%%%%%%%%%%%%%%%%%%%%

\subsection{Component Analysis}
\label{subsec:component_analysis}

We also demonstrate the cases when dropping partial attention modules or connections in comparison with the complete AF-net.
As an example, the person ReID on VIPeR dataset~\cite{gray2007evaluating} with six typical configurations are compared in Fig.~\ref{fig:AF-net_analysis}(a).
The orange bar shown in its bottom indicates the performance with the complete AF-net, while the yellow one is the M-net which is considered as the baseline model without the attention modules.
The rest four bars are configured as:

\noindent\textit{(1) Blue: na\"ive attention modules per branch.} In each branch of AF-net, a na\"ive attention module is applied to extract the attentive features $\tilde{\mathbf{F}}^{i,i}, i\in\Omega=\lbrace 1, 2, 3\rbrace$.

\noindent\textit{(2) Cyan: discarding the middle-level attention maps and attentive features.} We discard both the attention maps and attentive features of the block $2^{\text{nd}}$, \ie~prune the modules that produce $\tilde{\mathbf{F}}^{2,k}$ and $\tilde{\mathbf{F}}^{i, 2}, \forall i, k \in \{1, 2, 3\}$.

\noindent\textit{(3) Purple: pruning one branch.} It discards the first MDA module $\mathcal{F}(\alpha^1)$.

\noindent\textit{(4) Light purple: pruning two branches.} The first two MDA modules $\mathcal{F}(\alpha^1)$ and $\mathcal{F}(\alpha^2)$ are discarded.

The results clearly prove that either cutting down the number of MDA modules or connections within this module will pull down the performance, and it is reasonable that these attention components complement each other to generate the comprehensive feature representation and thus gain a higher accuracy.
Two examples with Top-5 identification results shown in Fig.~\ref{fig:AF-net_analysis}(b-c) further demonstrate the effectiveness and indispensability of each component of the entire AF-net.
The complete network is superior to both the multi-level na\"ive attention modules (Fig.~\ref{fig:AF-net_analysis}(b)) and the single MDA module (Fig.~\ref{fig:AF-net_analysis}(c)).

%=========================== tb:comparison with dataset =====================================
\newcolumntype{P}[1]{>{\centering\arraybackslash}p{#1}}
\newcolumntype{M}[1]{>{\centering\arraybackslash}m{#1}}
\begin{table}[t]
\center{}
\footnotesize{
\begin{tabular}{M{1.3cm}|M{1.7cm}M{1.7cm}|M{1.85cm}}
\hline
 & PETA & RAP & \textbf{PA-100K} \\
\hline
\hline
\# scene & - & 26 & \textbf{598} \\
\hline
\# sample & 19,000 & 41,585 & \textbf{100,000} \\
\hline
\# attribute & 61 (+4) & 69 (+3) & \textbf{26} \\
\hline
\# tracklet & - & -  & \textbf{18,206} \\
\hline
resolution & from $17\times39$ to $169\times365$ & from $36\times92$ to $344\times554$ & \textbf{from $50\times100$ to $758\times454$} \\
\hline
\end{tabular}
}
\vskip +0.1cm
\caption{Comparison of the proposed PA-$100$K dataset with existing datasets. The attribute number listed in the parentheses indicates the multi-class attributes while the one outside means the number of binary attributes.}
\label{tb:comparison_dataset}
\end{table}
%=========================== tb:comparison with dataset =====================================

\section{Pedestrian Attribute Recognition}
\label{sec:exp_attribute}

We evaluate our HP-net on two public datasets comparing the state-of-the-art methods. In addition, we further propose a new large-scale pedestrian attribute dataset PA-$100$K with larger scene diversities and amount of samples.

% {\color{red}
% In addition to the existing datasets, we propose a larger pedestrian attribute dataset PA-$100$K with increased scene diversity and amount of samples.
% %
% We evaluate our proposed HP-net on these three datasets in comparison with recent state-of-the-art methods.
% }

\subsection{PA-$100$K Dataset}
\label{sub:pa_100k_datasets}

%-------------------------------------------------------------------------
% \subsection{Datasets and Settings}
% \label{subsec:attribute_dataset}

\newcolumntype{P}[1]{>{\centering\arraybackslash}p{#1}}
\newcolumntype{M}[1]{>{\centering\arraybackslash}m{#1}}
\begin{table*}[ht]
\center{}
\footnotesize{
%\begin{tabular}{l||lllll||ll|lllll||ll|l||ll}
\begin{tabular}{M{0.7cm}||M{0.55cm}M{0.55cm}M{0.55cm}M{0.55cm}M{0.55cm}|M{0.55cm}M{0.55cm}||M{0.55cm}M{0.55cm}M{0.55cm}M{0.55cm}M{0.55cm}|M{0.55cm}M{0.55cm}||M{0.55cm}|M{0.55cm}M{0.55cm}}
\hline
Dataset         & \multicolumn{7}{c||}{RAP}                                  & \multicolumn{7}{c||}{PETA}                                 & \multicolumn{3}{c}{PA-100K}   \\
\hline
Method   & ELF-mm & FC7-mm & FC6-mm & ACN   & Deep-Mar & M-net & HP-net & ELF-mm & FC7-mm & FC6-mm & ACN   & Deep-Mar & M-net & HP-net & Deep-Mar & M-net & HP-net \\
\hline
mA     & 69.94  & 72.28  & 73.32  & 69.66 & 73.79   & 74.44    & \textbf{76.12}     & 75.21  & 76.65  & 77.96  & 81.15 & \textbf{82.6}    & 80.58    & 81.77     & 72.7    & 72.3     & \textbf{74.21}     \\
Accu   & 29.29  & 31.72  & 33.37  & 62.61 & 62.02   & 64.99    & \textbf{65.39}     & 43.68  & 45.41  & 48.13  & 73.66 & 75.07   & 75.68    & \textbf{76.13}     & 70.39   & 70.44    & \textbf{72.19}     \\
Prec  & 32.84  & 35.75  & 37.57  & \textbf{80.12} & 74.92   & 77.83    & 77.33     & 49.45  & 51.33  & 54.06  & 84.06 & 83.68   & 84.81    & \textbf{84.92}     & 82.24   & 81.7     & \textbf{82.97}     \\
Recall  & 71.18  & 71.78  & 73.23  & 72.26 & 76.21   & 77.89    & \textbf{78.79}     & 74.24  & 75.14  & 76.49  & 81.26 & 83.14   & 82.9     & \textbf{83.24}     & 80.42   & 81.05    & \textbf{82.09}     \\
F1     & 44.95  & 47.73  & 49.66  & 75.98 & 75.56   & 77.86    & \textbf{78.05}     & 59.36  & 61     & 63.35  & 82.64 & 83.41   & 83.85    & \textbf{84.07}     & 81.32   & 81.38    & \textbf{82.53}     \\
\hline
\end{tabular}
}
% \vskip +0.25cm
\caption{Quantitative results(\%) on three datasets for pedestrian attribute recognition, compared with previous benchmark methods.}
\label{tb:quant_result}
\end{table*}

Most of existing public pedestrian attribute datasets~\cite{deng2014pedestrian,li2016richly} only contain a limited number of scenes (at most $26$) with no more than $50,000$ annotated pedestrians.
To further evaluate the generality of the proposed method, we construct a new large-scale pedestrian attribute (PA) dataset named as PA-$100$K with $100,000$ pedestrian images from $598$ scenes, and therefore offer a superiorly comprehensive dataset for pedestrian attribute recognition.
To our best knowledge, it is to-date the largest dataset for pedestrian attribute recognition.
We compare our PA-100K dataset with the other two publicly available datasets in Table~\ref{tb:comparison_dataset}.

The samples of one person in \textbf{PETA dataset}~\cite{deng2014pedestrian} are only annotated once by randomly picking one exemplar image, and therefore share the same annotated attributes even though some of them might not be visible and some other attributes are ignored.
Another limitation is that the random partition of the training, validation and test sets are conducted in the whole dataset with no consideration of the person's identity across images, which leads to unfair image assignment of one person in different sets.
In \textbf{RAP dataset}~\cite{li2016richly}, the high-quality indoor images with controlled lighting conditions contain much lower variances than those under unconstrained real scenarios.
Moreover, some attributes are even highly imbalanced.

% \vspace{0.1cm}
% \noindent
\textbf{PA-$100$K dataset}
surpasses the the previous datasets both in quantity and diversity, as shown in Table~\ref{tb:comparison_dataset}.
We define $26$ commonly used attributes including global attributes like gender, age, and object level attributes like handbag, phone, upper-clothing and \etc.
The PA-$100$K dataset was constructed by images captured from real outdoor surveillance cameras which is more challenging. % and much closer to the real surveillance scenarios.
Different from the existing datasets, the images were collected by sampling the frames from the surveillance videos, which makes some future applications available, such as video-based attribute recognition and frame-level pedestrian quality estimation.
We annotated all pedestrians in each image and abandoned pedestrians with blurred motion or extreme low resolution (lower than $50\times100$).
The whole dataset is randomly split into training, validation and test sets with a ratio of $8:1:1$.
The samples of one person was extracted along its tracklets in a surveillance video, and they are randomly assigned to one of these sets, in which case PA-$100$K dataset ensures the attributes are learned independent of the person's identity.
All these sets are guaranteed to have positives and negatives of the $26$ attributes.
Note that this partition based on tracklets is fairer than the partition that randomly shuffles the images in PETA dataset.

In the following experiments, we employ five evaluation criteria\footnote{Criterion definitions are the same as those in ~\cite{li2016richly}.} including a label-based metric mean accuracy (mA), and four instance-based metrics, \ie~accuracy, precision, recall and F1-score.
To address the issue of imbalanced classes, we adopt a weighted cross-entropy loss function introduced by~\cite{li2015multi}.

\subsection{Comparison with the Prior Arts}
\label{subsec:attribute_comp_results}

We quantitatively and qualitatively compare the performance of the proposed method with the previous state-of-the-art methods on the previously mentioned three datasets.
The following comparisons keep the same settings as the prior arts on different datasets respectively.

%%%%%%%%%%%%%%%%%%%%% fig: result_pa100k %%%%%%%%%%%%%%%%%%%%%%%%%%%
\begin{figure}[t]
\centering
\includegraphics[width=1\linewidth]{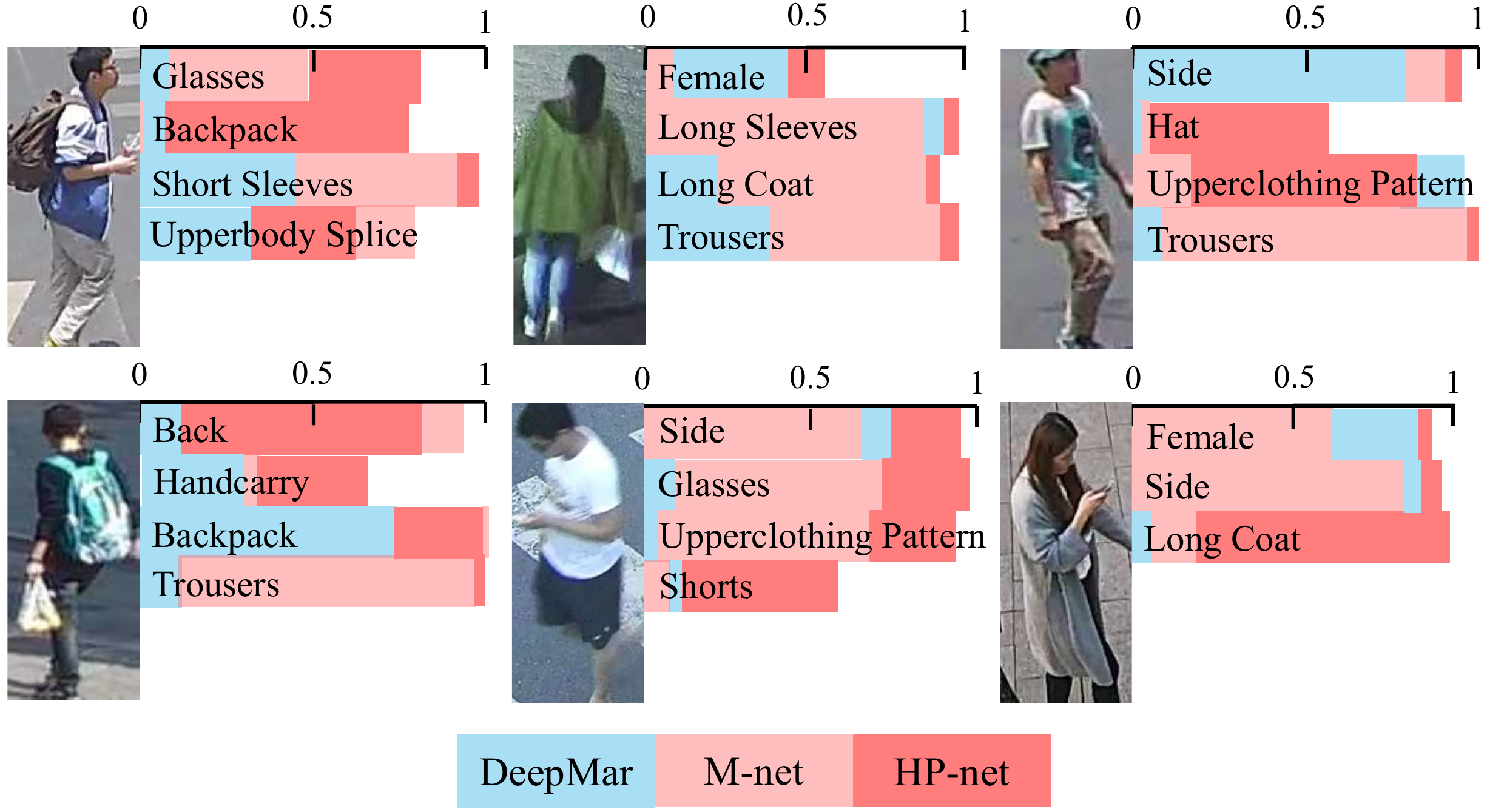}
\caption{
Comparison results between DeepMar, M-net and HP-net on partial ground truth attributes annotated for the given examples. Different colors represent different methods. Bars are plot by the prediction probabilities.
}
\label{fig:result_pa100k}
\vspace{-0.2cm}
\end{figure}
%%%%%%%%%%%%%%%%%%%%% fig: result_pa100k %%%%%%%%%%%%%%%%%%%%%%%%%%%

\noindent\textbf{Quantitative Evaluation.}
We list the results of each method on RAP, PETA and PA-$100$K datasets in Table~\ref{tb:quant_result}.
Six reference methods are selected to be compared with the proposed model.
The first three models are based on SVM classifier with hand-crafted features (ELF-mm~\cite{gray2008viewpoint,prosser2010person}) and deep-learned features (FC7-mm and FC6-mm) respectively.
% different features, where the Ensemble of Localized Features (ELF-mm~\cite{gray2008viewpoint,prosser2010person}) uses an ensemble of hand-crafted features, and FC7-mm as well as FC6-mm use CNN features extracted from BVLC Reference CaffeNet trained on ImageNet.
%
ACN~\cite{sudowe2015person} and DeepMar~\cite{li2015multi} are CNN models that achieved good performances by joint training the multiple attributes.

The baseline M-net and the proposed final model significantly outperform the state-of-the-art methods.
We are also interested in the performance of each attribute.
The bar in Fig.~\ref{fig:comp_bar_rap} shows the overlapped histograms of the mean accuracy (mA) for all attributes by DeepMar and HP-net.
The bars are sorted in descending order according to the larger mA between these methods at one attribute.
We find that the envelope superimposing the histogram is always supported by the HP-net with prominent performance gain against DeepMar, and is extremely superior on attributes which require fine-grained localization, like glasses and handbags.

% \vspace{+0.1cm}
\noindent\textbf{Qualitative Evaluation.}
Besides the quantitative results in Table~\ref{tb:quant_result}, we also conduct qualitative evaluations for exemplar pedestrian images.
As shown in the examples in Fig.~\ref{fig:comp_bar_rap}, sample images from RAP dataset and their attention maps demonstrate the localizability of the learned attention maps.
Especially in the first image, the attention map highlights two bags simultaneously.
We also notice a failure case on the attribute ``talking'' which is irrelevant to a certain region but requires a global understanding of the whole image.

For the PA-$100$K dataset, we show attribute recognition results for several exemplar pedestrian images in Fig.~\ref{fig:result_pa100k}.
The bars indicate the prediction probabilities.
Although the probabilities of one attribute do not directly imply its actual recognition confidences, they uncover the discriminative power of different methods as the lower probability corresponds to ambiguity or difficulty in correctly predicting one attribute.
The proposed HP-net reliably predicts these attributes with region-based saliency, like ``glasses'', ``back-pack'', ``hat'', ``shorts'' and ``handcarry''.

%-------------------------------------------------------------------------
%%%%%%%%%%%%%%%%%%%%% Experiment :: Person Re-identification %%%%%%%%%%%%%%%%%%%
% \subsection{Person Re-identification}
% \label{subsec:exp_reid}
\section{Person Re-identification}
\label{sec:exp_reid}

Referring to the person re-identification,
we also evaluate the HP-net with several reference methods on three publicly available datasets, quantitatively and qualitatively.

\begin{table}
\begin{center}
\begin{footnotesize}
\begin{tabular}{c|ccc}
\hline
Dataset & Market-1501~\cite{zheng2015scalable} & CUHK03~\cite{li2014deepreid} & VIPeR~\cite{gray2007evaluating} \\\hline
\# identities & 1501 & 1360 & 632\\
\# images & 32643 & 13164 & 1264\\
\# cameras & 6 & 2 & 2\\\hline
\# training IDs & 750 & 1160 & 316\\\hline
\# test IDs & 751 & 100 & 316\\
\# probe images & 3368 & 100 & 316\\
\# gallery images & 19732 & 100 & 316\\\hline
\end{tabular}
\end{footnotesize}
% \vspace{+0.1cm}
\caption{The specifications of three evaluated ReID datasets.}
\label{tab:datasets}
\vspace{-0.3cm}
\end{center}
\end{table}

\subsection{Datasets and Setups}

The proposed approach is evaluated on three publicly standard datasets, including
CUHK03~\cite{li2014deepreid}, VIPeR~\cite{gray2007evaluating}, and Market-1501~\cite{zheng2015scalable}.
A summary about the statistical information of the three datasets are listed in Table~\ref{tab:datasets}.
%CUHK03 [23] is one of the most largest published person re-identification datasets, it consists of five different pairs of camera views, and has more than 14,000 images of 1467 pedestrians.
%PRID [15] extracts pedestrian images from recorded trajectory video frames. It has two camera views, each contains 385 and 749 identities, respectively. But only 200 of them appear in both views.
%VIPeR [13] is one of the most challenging dataset, since it has 632 people but with various poses, viewpoints, image resolutions, and lighting conditions.
For the Market-1501 dataset, the same data separation strategy is used as~\cite{zheng2015scalable}.
For the other datasets, the training, validation and testing images are sampled based on the strategy introduced in~\cite{xiao2016learning}.
The training and validation identities are guaranteed to have no overlaps with the testing ones for all evaluated datasets.
Following the pipeline of JSTL~\cite{xiao2016learning}, all the training samples are combined together to train a single ReID model from scratch, which can be directly evaluated on all the testing datasets.

The widely applied cumulative match curve (CMC) metric is adopted for quantitative evaluation.
While in the matching process, the cosine distance is computed between each query image and all the gallery images, and the ranked gallery list is returned.
All the experiments are conducted under the setting of single query and the testing procedure is repeated 100 times to get an average result.

\begin{table}[t]
\begin{footnotesize}
\begin{center}
\begin{tabular}{c|cccc}
\hline
\textbf{CUHK03}&\textbf{Top-1}&\textbf{Top-5}&\textbf{Top-10}&\textbf{Top-20}\\\hline
PersonNet~\cite{wu2016personnet} & 64.8 & 89.4 & 94.9 & 98.2 \\
JSTL~\cite{xiao2016learning} & 75.3 & - & - & - \\
Joint ReID~\cite{ahmed2015improved} & 54.7 & - & - & - \\
LOMO-XQDA~\cite{liao2015person} & 52.2 & - & - & - \\\hline
M-net & \textbf{88.2} & 98.2 & 99.1 & 99.5\\
HP-net  & \textbf{91.8} & 98.4 & 99.1 & 99.6\\
\hline
\hline
\textbf{VIPeR}&\textbf{Top-1}&\textbf{Top-5}&\textbf{Top-10}&\textbf{Top-20}\\\hline
NFST~\cite{zhang2016learning} & 51.2 & 82.1 & 90.5 & 96.0\\
SCSP~\cite{chen2016similarity}& \textbf{53.5} & 82.6 & 91.5 & 96.7\\
GOG+XQDA~\cite{matsukawa2016hierarchical} & 49.7 & 79.7 & 88.7 & 94.5 \\
TCP~\cite{cheng2016person} & 47.8 & 74.7 & 84.8 & 91.1 \\\hline
M-net & 51.6 & 73.1 & 81.6 & 88.3\\
HP-net & \textbf{56.6} & 78.8 & 87.0 & 92.4\\
\hline
\hline
\textbf{Market-1501}&\textbf{Top-1}&\textbf{Top-5}&\textbf{Top-10}&\textbf{Top-20}\\\hline
WARCA-L~\cite{jose2016scalable} & 45.2 & 68.1 & 76.0 & 84.0 \\
LOMO+CN~\cite{varior2016siamese} & 61.6 & - & - & - \\
S-CNN~\cite{varior2016gated} & \textbf{65.9} & - & - & - \\
BoW-best~\cite{zheng2015scalable} & 44.4 & 63.9 & 72.2 & 79.0 \\\hline
M-net & 73.1 & 89.5 & 93.4 & 96.0 \\
HP-net & \textbf{76.9} & 91.3 & 94.5 & 96.7 \\
\hline
\end{tabular}
\end{center}
\caption{Experimental results(\%) of the proposed HP-net and other comparisons on three datasets. The CMC Top-1-5-10-20 accuracies are reported. The Top-1 accuracies of two best performing approaches are marked in bold.}
\label{tab:accuracy}
\vspace{-0.1cm}
\end{footnotesize}
\end{table}

\subsection{Performance Comparisons}

\noindent\textbf{Quantitative Evaluation.}
As shown in Table~\ref{tab:accuracy}, the proposed approach is compared with a series of the deep neural networks like PersonNet~\cite{wu2016personnet}, the multi-domain CNN JSTL~\cite{xiao2016learning},
the Joint ReID method~\cite{ahmed2015improved}, and the horizontal occurrence model LOMO-XQDA~\cite{liao2015person} on CUHK03~\cite{li2014deepreid}.
As for the VIPeR~\cite{gray2007evaluating} dataset, the null space semi-supervised learning method NFST~\cite{zhang2016learning}, the similarity learning method SCSP~\cite{chen2016similarity}, the hierarchical Gaussian model GOG+XQDA~\cite{matsukawa2016hierarchical},
and the triplet loss model TCP~\cite{cheng2016person} are selected for comparison.
The Market-1501~\cite{zheng2015scalable} dataset is also evaluated with the metric learning WARCA-L~\cite{jose2016scalable}, a novel Siamese LSTM architecture LOMO+CN~\cite{varior2016siamese}, the Siamese CNN with learnable gate S-CNN~\cite{varior2016gated},
and the bag of words model BoW-best~\cite{zheng2015scalable}.

Besides the results of the proposed approach with complete HP-net, the results of the M-net are also listed as the baseline for the three datasets.
From Table~\ref{tab:accuracy}, we can observe that the proposed approach achieves the Top-1 accuracies of $91.8\%$, $56.6\%$ and $76.9\%$ on the CUHK03, ViPeR and Market-1501 datasets, respectively, and it achieves the state-of-the-art performance on all the three datasets.
Moreover, even though the M-net can achieve quite satisfactory results on all datasets, the proposed pipeline can further improve the Top-1 accuracies by $3.6\%$, $5.0\%$, and $3.8\%$ for each dataset, respectively.
%
% It demonstrates that the proposed framework can help to learn more discriminative features even from  baseline.

\begin{figure}[t]
\small
\centering
\includegraphics[width=1\linewidth]{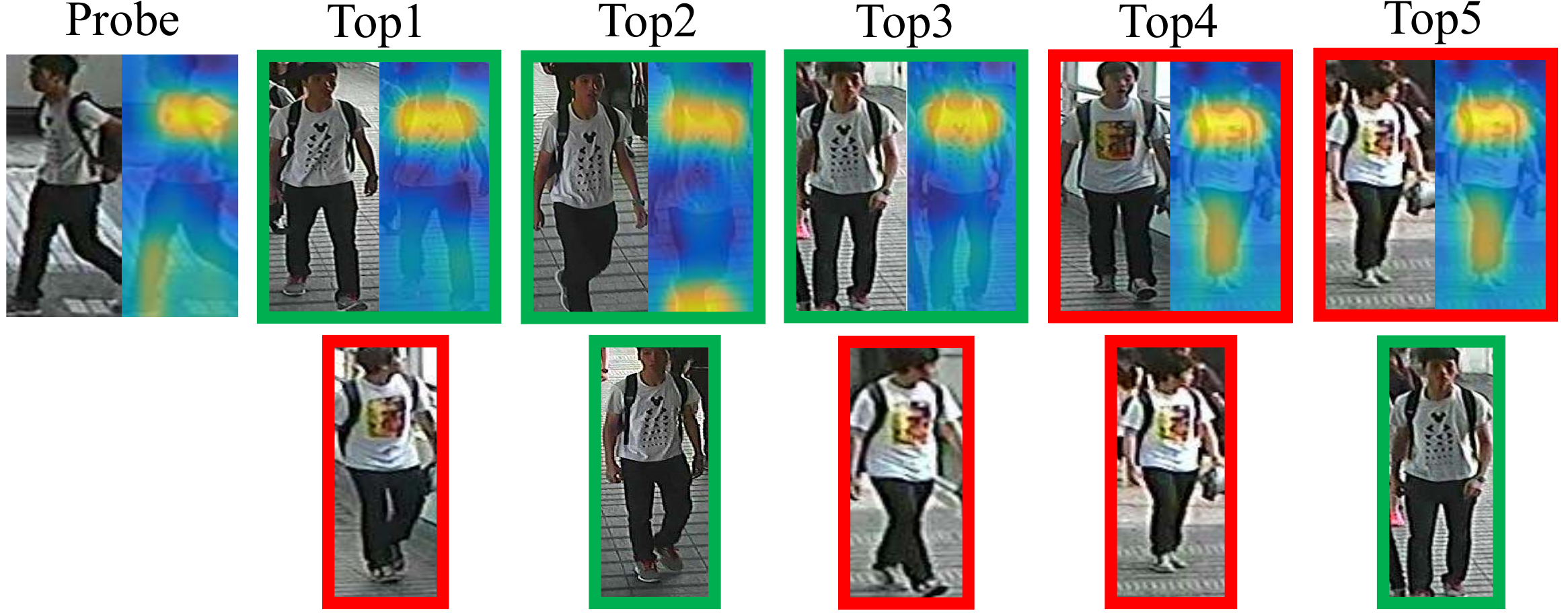}
\caption{
Comparison results between HP-net and M-net. For the probe images, the Top-5 retrieval results of HP-net together with attention maps are shown in the first row, and the results of M-net are shown in the second row.
}
\label{fig:result_reid}
% \vspace{-0.1cm}
\end{figure}

\vspace{+0.1cm}
\noindent\textbf{Qualitative Evaluation.}
To highlight the performance of the proposed method on extracting localized semantic features, one query image together with its Top-5 gallery results by the proposed method and the M-net are visualized in Fig.~\ref{fig:result_reid}.
We observe that the proposed approach improves the rankings of the M-net and gets the correct results.
%
% By visualizing the attention maps of the query image and the Top-5 gallery images of both methods, we observe that the proposed attention modules can successfully locate the T-shirt patterns, in which the fine-grained features are extracted and discriminatingly identify the query person against the other identities with similar dressing.
By visualizing the attention maps from HP-Net of the query images and the Top-5 gallery images of both methods, we observe that the proposed attention modules can successfully locate the T-shirt patterns, in which the fine-grained features are extracted and discriminatingly identify the query person against the other identities with similar dressing.
\

%%%%%%%%%%%%%%%%%%%%% fig: result_reid %%%%%%%%%%%%%%%%%%%%%%%%%%%
%%%%%%%%%%%%%%%%%%%%% fig: result_reid %%%%%%%%%%%%%%%%%%%%%%%%%%%

\section{Conclusion}
\label{sec:conclusion}
\vspace{-0.1cm}
In this paper, we present a new deep architecture called HydraPlus network with a novel multi-directional attention mechanism.
Extensive ablation studies and experimental evaluations have manifested the effectiveness of the HP-net to learn multi-level and multi-scale attentive feature representations for fine-grained tasks in pedestrian analysis, like pedestrian attribute recognition and person re-identification.
In the end, a new large-scale attribute dataset PA-$100$K is introduced to facilitate various pedestrian analysis tasks.

\vspace{+5pt}
\noindent\textbf{Acknowledgement} This work is supported in part by SenseTime Group Limited, in part by the General Research Fund through the Research Grants Council of Hong Kong under Grants CUHK14213616, CUHK14206114, CUHK14205615, CUHK419412, CUHK14203015, CUHK14207814, and in part by the Hong Kong Innovation and Technology Support Programme Grant ITS/121/15FX.

{\small
\bibliographystyle{ieee}
\bibliography{related_work_v3}
}

\end{document}